\documentclass[runningheads]{llncs}

\usepackage{eccv}

\usepackage{eccvabbrv}

\usepackage{graphicx}
\usepackage{booktabs}

\usepackage[accsupp]{axessibility}  

\usepackage{hyperref}

\usepackage{orcidlink}
\usepackage{float}
\usepackage{subcaption}
\usepackage{multirow}
\usepackage{makecell}

\usepackage{appendix}
\makeatletter
\newcommand{\@chapapp}{\relax}%
\makeatother

\newcommand{\ours}{SpaRP\xspace}
\newcommand{\suppmat}{Appendix\xspace}

\makeatletter
\newcommand{\myfnsymbol}[1]{%
  \expandafter\@myfnsymbol\csname c@#1\endcsname
}
\newcommand{\@myfnsymbol}[1]{%
  \ifcase #1
  \or 1
  \or 2
  \or \TextOrMath{\textasteriskcentered}{*}
  \or \TextOrMath{\textdagger}{\dagger}
  \or $\ddagger$ 
  \fi
}
\newcommand{\affiliationA}{\@myfnsymbol{1}}
\newcommand{\affiliationB}{\@myfnsymbol{2}}
\newcommand{\equalcontributor}{\@myfnsymbol{3}}
\newcommand{\intern}{\@myfnsymbol{4}}
\newcommand{\corresponding}{\@myfnsymbol{5}}
\makeatother

\begin{document}

\title{\ours: Fast 3D Object Reconstruction and \\ Pose Estimation from Sparse Views \vspace{-0.7em}} 

\titlerunning{\ours}

\author{Chao Xu\inst{1,2}\textsuperscript{\intern}\orcidlink{0009-0001-0574-5357} \and
Ang Li\inst{3} \and
Linghao Chen\inst{2,4}\textsuperscript{\intern} \and 
Yulin Liu\inst{5} \and \\
Ruoxi Shi\inst{2,5}\textsuperscript{\intern} \and
Hao Su\inst{2,5}\textsuperscript{\corresponding} \and
Minghua Liu\inst{2,5}\textsuperscript{\intern\corresponding}}

\authorrunning{Xu et al.}

\institute{UCLA \and
Hillbot Inc. \and
Stanford University \\ \and
Zhejiang University \and 
UC San Diego \\}

\renewcommand{\thefootnote}{\myfnsymbol{footnote}}
\maketitle
\footnotetext[4]{This work is done while the author is an intern at Hillbot Inc. \textsuperscript{$\ddagger$} Equal advisory.}

\setcounter{footnote}{0}
\renewcommand{\thefootnote}{\fnsymbol{footnote}}

\vspace{-2em}
\begin{figure}[H]
  \centering
  \includegraphics[width=\linewidth]{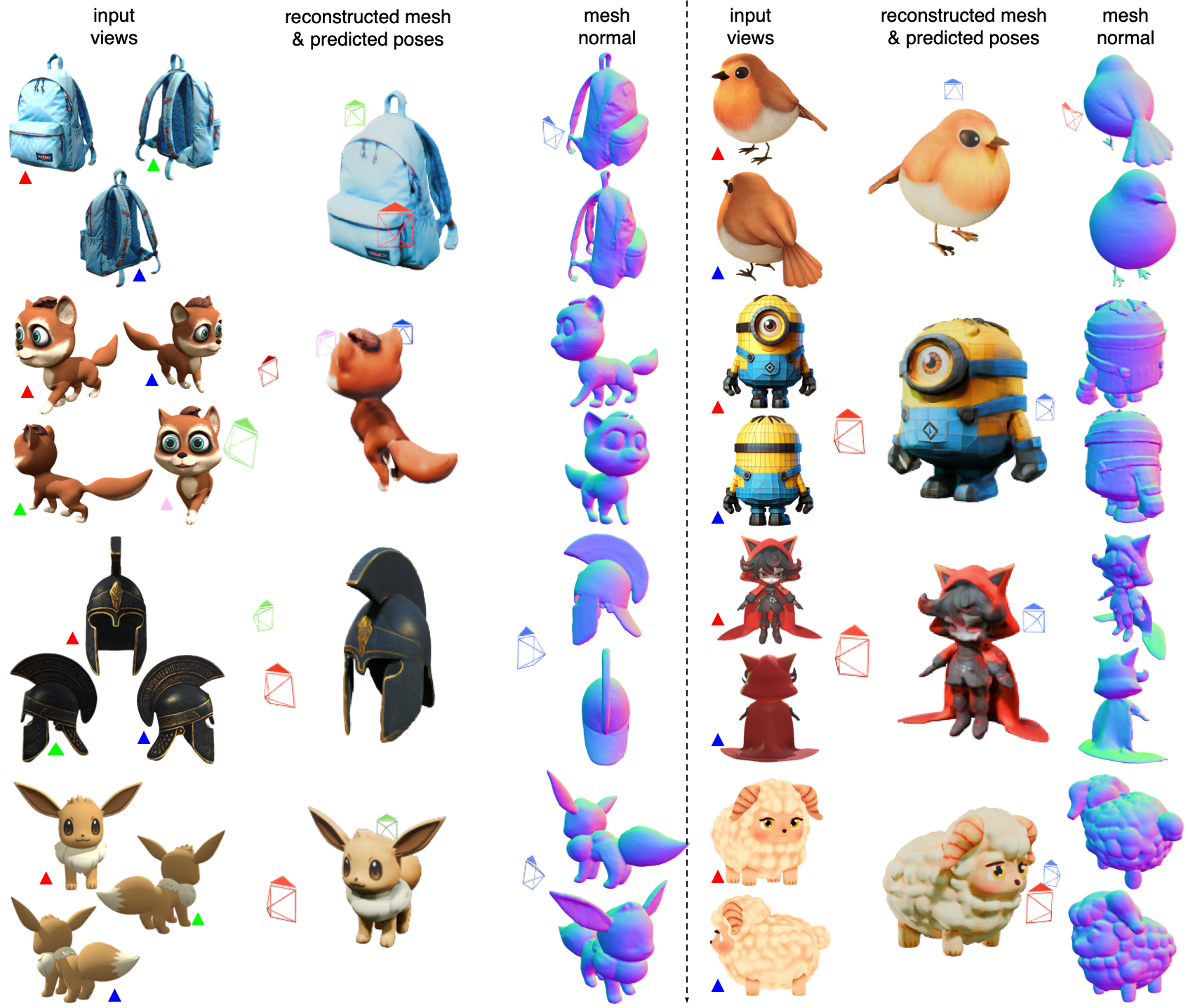}
   \vspace{-2em}
  \caption{\ours handles open-world 3D reconstruction and pose estimation from unposed sparse-view images, delivering results in approximately 20 seconds.}
  \label{fig:teaser}
\end{figure}

\begin{abstract}
\vspace{-1em}
Open-world 3D generation has recently attracted considerable attention. While many single-image-to-3D methods have yielded visually appealing outcomes, they often lack sufficient controllability and tend to produce hallucinated regions that may not align with users' expectations. In this paper, we explore an important scenario in which the input consists of one or a few unposed 2D images of a single object, with little or no overlap. We propose a novel method, \ours, to reconstruct a 3D textured mesh and estimate the relative camera poses for these sparse-view images. \ours distills knowledge from 2D diffusion models and finetunes them to implicitly deduce the 3D spatial relationships between the sparse views. The diffusion model is trained to jointly predict surrogate representations for camera poses and multi-view images of the object under known poses, integrating all information from the input sparse views. These predictions are then leveraged to accomplish 3D reconstruction and pose estimation, and the reconstructed 3D model can be used to further refine the camera poses of input views. Through extensive experiments on three datasets, we demonstrate that our method not only significantly outperforms baseline methods in terms of 3D reconstruction quality and pose prediction accuracy but also exhibits strong efficiency. It requires only about 20 seconds to produce a textured mesh and camera poses for the input views. Project page: \url{https://chaoxu.xyz/sparp}.
   \vspace{-0.5em}
\end{abstract}
 \vspace{-2.7em}
\section{Introduction}
\label{sec:intro}
 \vspace{-0.7em}
3D object reconstruction is a long-standing problem with applications spanning 3D content creation, augmented reality, virtual reality, and robotics, among others. Although traditional photogrammetry~\cite{barnes2009patchmatch,schonberger2016pixelwise,stereopsis2010accurate} and recent neural field methods~\cite{mildenhall2021nerf,wang2021neus,yariv2021volume} have made significant strides in reconstructing high-fidelity geometry and appearance, they typically require dense view inputs. However, in many practical scenarios, such as in e-commerce and consumer capture situations, acquiring a comprehensive set of high-resolution images along with precise camera data is not always feasible.

On the other end of the spectrum, the tasks of converting a single image to 3D and text to 3D have recently seen substantial progress~\cite{poole2022dreamfusion,lin2023magic3d,li2023instant3d,liu2023one2345++,shi2023mvdream,wang2023prolificdreamer}, thanks to the rich priors embedded in 2D diffusion models~\cite{ramesh2021zero,saharia2022photorealistic,rombach2022high} and pre-training on extensive 3D datasets~\cite{deitke2023objaverse}. These methods may achieve high-quality geometry and texture that matches the input view, but they also introduce ambiguities in the regions not visible in the input image (such as the back view). Although these methods attempt to hallucinate reasonable interpretations of these invisible areas, the generated regions may not always align with users' expectations, and users often lack sufficient control over these ambiguous regions.

In this paper, we explore a critical scenario where the input consists of one or a few unposed 2D images of a single object. The images are captured from arbitrarily distributed camera poses, often with little to no overlap. We tackle both the 3D reconstruction and pose estimation of input images under this sparse view setting. Note that, in dense view setting, traditional Structure-from-Motion (SfM) solvers (e.g., COLMAP~\cite{schoenberger2016sfm}) are typically employed for pose estimation. However, with sparse view inputs, these solvers often become unreliable and tend to fail due to insufficient overlapping visual cues. This issue is the main reason why existing sparse view reconstruction methods~\cite{kim2022infonerf,long2022sparseneus,zhou2023sparsefusion} generally require known camera poses as input. While some recent methods have attempted pose-free reconstruction and pose estimation for sparse views~\cite{lin2023relpose++,sinha2023sparsepose,zhang2022relpose,jiang2022few,jiang2023leap}, they are usually trained on a predefined small set of object categories and exhibit poor generalization to unseen object categories.

In response, we propose an innovative class-agnostic approach called \ours, capable of processing arbitrary object categories with unposed sparse views. Our inspiration comes from recent breakthroughs in open-domain single-image-to-3D methods. They leverage 2D diffusion models (e.g., Stable Diffusion~\cite{rombach2022high}) to generate novel viewpoints of an object~\cite{liu2023zero}, and even consistent multi-view images from a single input image~\cite{shi2023mvdream,shi2023zero123++,li2023instant3d,liu2023syncdreamer,long2023wonder3d}, by finetuning the diffusion models with corresponding multi-view image pairs. These discoveries imply that 2D diffusion models harbor rich priors concerning 3D objects. Instead of merely producing multi-view images, we contemplate leveraging 2D diffusion models to examine a set of unposed input images from sparse viewpoints, infer their spatial interrelationships, and recover relative camera poses and underlying 3D shapes.

Specifically, we finetune a 2D diffusion model~\cite{rombach2022high} to process sparse input views by compositing them into a single image for conditioning. The diffusion model is concurrently tuned to deduce the relative poses of the input images and the underlying 3D objects. For the relative pose estimation branch, instead of outputting camera poses as scalars, we task 2D diffusion models to produce a surrogate representation: the NOCS maps~\cite{wang2019normalized} that embed pixel-wise correspondences across different views and are more suitable for 2D diffusion models. From these maps, we extract the relative camera poses for the sparse views using the traditional PnP algorithm~\cite{opencvpnp}, assuming known camera intrinsics. For the reconstruction branch, the diffusion model is tasked to produce multi-view images of the object from fixed known camera poses, covering the entire 3D object. This task requires the models to incorporate all information from input sparse views and hallucinate invisible regions. We then feed the generated images with fixed known poses into a pre-trained 3D reconstruction module~\cite{liu2023one2345++} to create a textured 3D mesh. We can further refine the estimated camera poses by aligning the input views with the generated mesh through differentiable rendering~\cite{Laine2020diffrast}.

We train \ours on the Objaverse~\cite{deitke2023objaverse} dataset with 1–6 unposed input views. Unlike some previous methods that rely on costly per-shape optimization~\cite{wu2023ifusion}, our method delivers 3D textured meshes along with camera poses in a much more efficient manner, requiring only $\sim$16 seconds. As shown in \cref{fig:teaser}, our approach can faithfully generate 3D assets that closely follow the reference unposed images, effectively overcoming the ambiguity issue of single-image-to-3D. Extensive evaluation on three datasets demonstrates the superior performance of our method over baselines in reconstructing 3D meshes with vivid appearance and high-fidelity geometry, alongside precise pose estimation of the input images.

\vspace{-1.em}
\section{Related Work}
\label{sec:relatedWork}
\vspace{-.5em}

\subsection{Sparse-View 3D Reconstruction}

Reconstructing 3D objects from sparse-view images is challenging due to the lack of visual correspondence and clues. When a small baseline between images is assumed, several methods~\cite{chen2021mvsnerf,long2022sparseneus,rematas2021sharf,trevithick2021grf,yu2021pixelnerf,johari2022geonerf,kulhanek2022viewformer,liu2022neural,ren2023volrecon,wang2022attention,wang2021ibrnet,yang2023contranerf} have pretrained generalizable models to infer surface positions by establishing pixel correspondences and learning generalizable priors across scenes. However, these methods often fail to produce satisfactory results when the sparse-view images have a large baseline. Some studies have attempted to alleviate the dependence on dense views by incorporating priors or adding regularization~\cite{shi2023zerorf,jain2021putting,kim2022infonerf,niemeyer2022regnerf} into the NeRF optimization process. Others have employed 2D diffusion priors to generate novel-view images as additional input for the NeRF model~\cite{zhou2023sparsefusion,chan2023genvs,tewari2023diffusion,karnewar2023holodiffusion}. For example, ReconFusion~\cite{wu2023reconfusion} trains a NeRF from sparse-view images and uses a denoising UNet to infer some novel view images as support for the NeRF model. EscherNet~\cite{kong2024eschernet} utilizes Stable Diffusion for novel view synthesis and designs a camera positional encoding module to yield more consistent images. Furthermore, some recent works~\cite{long2023wonder3d,liu2023syncdreamer,shi2023mvdream} have integrated specialized loss functions and additional modalities as inputs into NeRF-based per-scene optimization.

In contrast to these methods, our approach does not require camera poses for the input sparse views. It is not limited to small baselines and is capable of generating 360-degree meshes. Furthermore, without the need for per-shape optimization, our method can quickly produce both textured meshes and camera poses in about 20 seconds.

\vspace{-.5em}
\subsection{Pose-Free Reconstruction}
\vspace{-.2em}
 
Unlike the methods mentioned above, which assume known camera poses, many studies have aimed to solve the pose-free reconstruction challenge. When provided with dense images, some approaches ~\cite{lin2021barf,wang2021nerf,xia2022sinerf} jointly optimize the NeRF representation along with camera parameters. However, due to the highly non-convex nature of this optimization problem, such methods are susceptible to initial pose guesses and can become trapped in local minima. This issue worsens when input images are sparse, with increasing ambiguity and reduced constraint availability. In response, numerous proposals have attempted to enhance optimization robustness. For example, SpaRF~\cite{truong2023sparf} uses dense image matches as explicit optimization constraints, while FvOR~\cite{yang2022fvor} starts with coarse predictions of camera poses and alternated updates between shape and pose.

In contrast to the optimization-based methods, there is a body of research proposing generalizable solutions for this problem. VideoAE~\cite{lai2021video} infers scene geometry from the first frame in a video series and estimates camera poses relative to that frame, which allows for warping scene geometry to decode new viewpoints. SparsePose~\cite{sinha2023sparsepose} first regresses and then iteratively refines camera poses. FORGE~\cite{jiang2022few} designs neural networks to infer initial camera poses, fuse multi-view features, and decode spatial densities and colors. GRNN~\cite{tung2019learning} offers a GRU-based reconstruction method estimating the relative pose for each input view against a global feature volume. The RelPose series~\cite{zhang2022relpose,lin2023relpose++} use probabilistic modeling for relative rotation estimation between images. Other works~\cite{sajjadi2022scene,jiang2023leap} eschew explicit camera pose estimations, instead employing transformers to encode input views into latent scene representations for novel view synthesis.

More recently, leveraging large vision models and diffusion models, which have shown significant promise, new efforts have emerged for camera pose estimation. PoseDiffusion~\cite{wang2023posediffusion} implements a diffusion model guided by 2D keypoint matches to estimate poses. PF-LRM~\cite{wang2023pf} adapts the LRM model~\cite{hong2023lrm} to predict a point cloud for each input image, then utilizes differentiable PnP for pose estimation. iFusion~\cite{wu2023ifusion} employs an optimization pipeline to assess relative elevations and azimuths. It utilizes Zero123~\cite{liu2023zero} predictions as a basis and optimizes the relative pose between two images by minimizing the reconstruction loss between the predicted and target images.

In contrast to these existing approaches, our proposal capitalizes on the extensive priors inherent in pre-trained 2D diffusion models, thereby providing exceptional generalizability to handle a diverse range of open-world categories. Our method predicts camera poses and 3D mesh geometry in a single feedforward pass, negating the need for per-shape optimization.

\vspace{-.5em}
\subsection{Open-World 3D Generation}
\vspace{-.2em}

Open-world single-image-to-3D and text-to-3D tasks have recently undergone significant advancements. Recent 2D generative models~\cite{ramesh2021zero,saharia2022photorealistic,rombach2022high} and vision-language models~\cite{radford2021learning} have supplied valuable priors about the 3D world, sparking a surge in research on 3D generation. Notably, models such as DreamFusion~\cite{poole2022dreamfusion}, Magic3D~\cite{lin2023magic3d}, and ProlificDreamer~\cite{wang2023prolificdreamer} have pioneered a line of approach to per-shape optimization~\cite{jain2022zero,melas2023realfusion,deng2023nerdi,mohammad2022clip,lee2022understanding,metzer2023latent,michel2022text2mesh,raj2023dreambooth3d,seo2023let,tang2023make,wang2023score,xu2023neurallift,xu2023dream3d,tang2023dreamgaussian,chen2023fantasia3d,qian2023magic123,chen2023text,yu2023points}. These models optimize a 3D representation (e.g., NeRF) for each unique text or image input, utilizing the 2D prior models for gradient guidance. Although they produce impressive results, these methods are hampered by prolonged optimization times, often extending to several hours, and ``multi-face issue'' problems.

Moreover, beyond optimization-based methods, exemplified by Zero123~\cite{liu2023zero}, numerous recent studies have investigated the employment of pre-trained 2D diffusion models for synthesizing novel views from single images or text~\cite{shi2023mvdream,liu2023syncdreamer,weng2023consistent123,ye2023consistent,long2023wonder3d,shi2023zero123++}. They have introduced varied strategies to foster 3D-consistent multi-view generation. The resulting multi-view images can then serve for 3D reconstruction, utilizing either optimization-based methods ~\cite{shi2023mvdream,liu2023syncdreamer,long2023wonder3d} or feedforward models~\cite{liu2023one2345,liu2023one2345++,li2023instant3d}.

While most existing works focus on single-image-to-3D or text-to-3D, they often hallucinate regions that are invisible in the input image, which provides users with limited control over those areas. In this paper, we seek to broaden the input to encompass unposed sparse views and address both the 3D reconstruction and pose estimation challenges in a time-efficient way—within tens of seconds.

\vspace{-1.em}
\section{Method}
\label{sec:method}
\vspace{-.5em}

\begin{figure}[!t]
    \centering
     \includegraphics[width=1\linewidth]{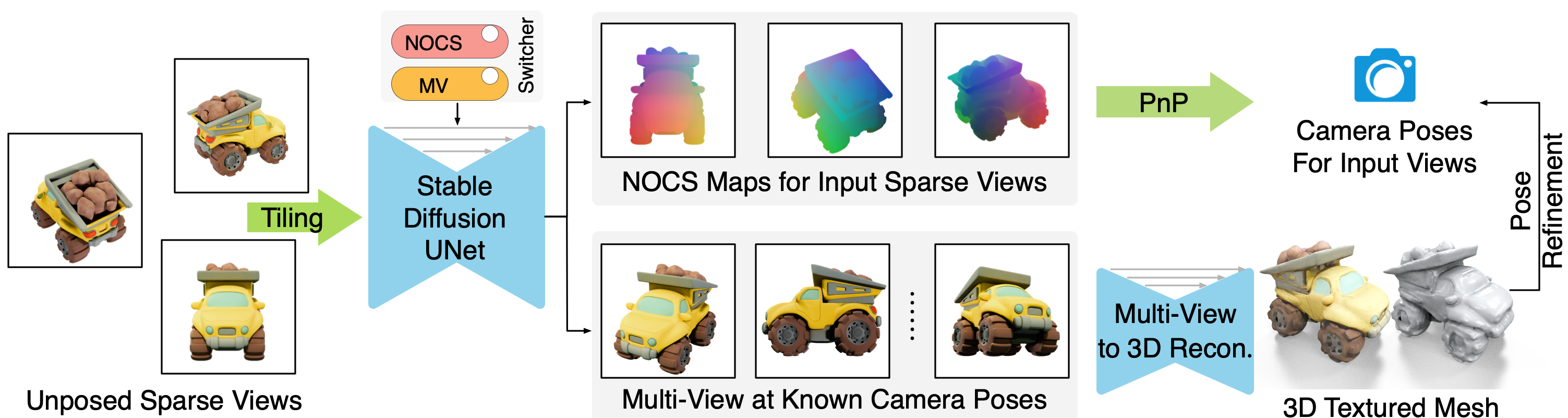}
    \vspace{-1.7em}
    \caption{\textbf{Pipeline Overview of \ours.} We begin by taking a sparse set of unposed images as input, which we tile into a single composite image. This composite image is subsequently provided to the Stable Diffusion UNet to serve as the conditioning input. The 2D diffusion model is simultaneously finetuned to predict NOCS maps for the input sparse views and multi-view images under known camera poses. From the NOCS maps, we extract the camera poses corresponding to the input views. The multi-view images are then processed by a reconstruction module to generate textured 3D meshes. Optionally, the camera poses can be further refined using the generated mesh for improved accuracy.}
    \vspace{-1.5em}
    \label{fig:method}
\end{figure}

Given $n$ unposed input images $\{ \mathbf{I}_i \mid i=1,\dots,n; \ 1\leq n \leq 6\}$, which illustrate a single object from arbitrary categories, we predict their relative camera poses ${\boldsymbol{\xi}_{ij}}$ and reconstruct the 3D model $\mathcal{M}$ of the object. As illustrated in \cref{fig:method}, we first finetune a 2D diffusion model~\cite{rombach2022high} to process the unposed sparse input images (\cref{sec:tiling}). The 2D diffusion model is responsible for jointly generating grid images for both the NOCS maps of the input views, as well as multi-view images with known camera poses. We use the predicted NOCS maps to estimate the camera poses for the input views (\cref{sec:nocs2pose}). The resulting multi-view images are fed into a two-stage 3D diffusion model for a coarse-to-fine generation of a 3D textured mesh (\cref{sec:recon}). This joint training strategy allows the two branches to complement each other. It enhances the understanding of both the input sparse views and the intrinsic properties of the 3D objects, thereby improving the performance of both pose estimation and 3D reconstruction. Optionally, the generated 3D mesh can also be used to further refine the camera poses (\cref{sec:pose_refine}).

\begin{figure}[t]
    \centering
     \includegraphics[width=1\linewidth]{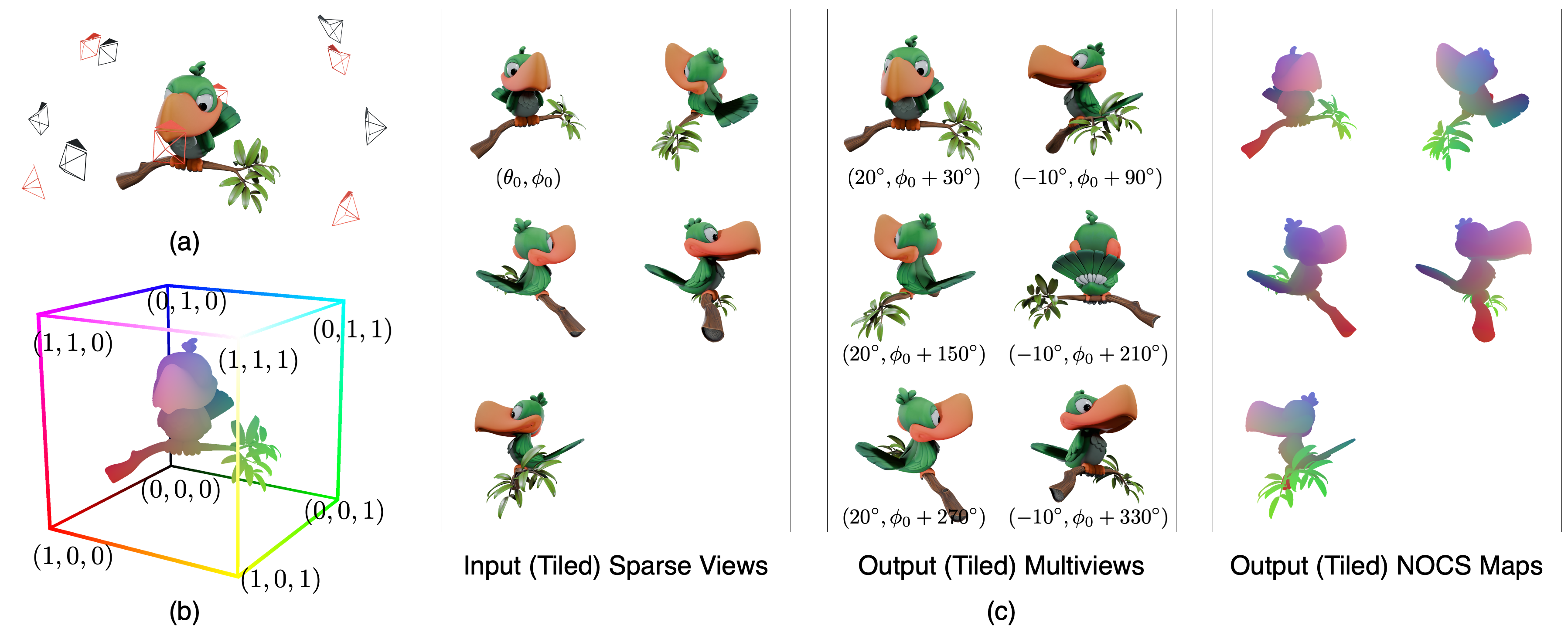}
    \vspace{-2em}
    \caption{(a) Regardless of the poses of the sparse input views (in black), the output multiviews are uniformly distributed (in red) and encompass the entire 3D object. (b) The Normalized Object Coordinate Space (NOCS) of the object, whose orientation is aligned with the azimuth of the first input view. (c) An example of input and output tiled images. The elevation and azimuth of the first input view are denoted by $\theta_0$ and $\phi_0$, respectively. The camera poses of the output multiview images are determined by $\phi_0$. The output NOCS maps correspond to the input sparse views, and the orientation of the coordinate frame is also determined by $\phi_0$.} 
    \vspace{-1.7em}
    \label{fig:inputoutput}
\end{figure}

\vspace{-.5em}
\subsection{Tiling Sparse View Images as Input Condition}
\vspace{-.2em}

\label{sec:tiling}

Recently, numerous studies have shown that 2D diffusion models not only possess robust open-world capabilities but also learn rich 3D geometric priors. For instance, Stable Diffusion~\cite{rombach2022high}, can be finetuned to include camera view control~\cite{liu2023zero,shi2023mvdream,shi2023zero123++,li2023instant3d,liu2023syncdreamer,long2023wonder3d}, enabling it to predict novel views of objects—a task that necessitates significant 3D spatial reasoning. Consequently, we are inspired to utilize the rich priors inherent in 2D diffusion models for the tasks of sparse view 3D reconstruction and pose estimation.

Unlike most existing approaches that use a single RGB image as the condition and focus on synthesizing multi-view images, our goal is to take a sparse set of input images and stimulate Stable Diffusion to infer the spatial relationships among the input views implicitly. To accomplish this, given $1 \sim 6$ sparse views from arbitrary camera poses, we tile them into a $3 \times 2$ multi-view grid, as illustrated in Fig.~\ref{fig:inputoutput} (c). The image in the first grid cell determines a canonical frame (to be discussed later), while the order of the other views is inconsequential. When there are fewer than 6 sparse views, we use empty padding for the remaining grid cells. This composite image then serves as the condition for Stable Diffusion, which is expected to assimilate all information from the input sparse views during the diffusion process.

We employ Stable Diffusion 2.1 as our base model. To adapt the original text-conditioning to our tiled multi-view image condition, we follow~\cite{shi2023zero123++} to apply both local and global conditioning strategies. For local conditioning, we use the reference-only attention mechanism~\cite{refattn}, where we process the reference tiled image with the denoising UNet model and append the attention keys and values from this image to corresponding layers in the denoising model for the target images. This mechanism facilitates implicit yet effective interactions between the diffusion model and the sparse views. For global conditioning, we integrate the mean-pooled CLIP embedding of all input images—modulated by learnable token weights—into the diffusion process, enhancing the model's ability to grasp the overarching semantics and structure of the sparse views.

As depicted in~\cref{fig:method,fig:inputoutput}, our objective is to concurrently generate grid images for both NOCS maps of the input views and multi-view images from known camera poses. To achieve this, we utilize a domain switcher~\cite{long2023wonder3d} that enables flexible toggling between the two domains. The switcher consists of two learnable embeddings, one for each domain, which are then injected into the UNet of the stable diffusion models by being added to its time embedding.

\vspace{-.5em}
\subsection{Image-to-NOCS Diffusion as a Pose Estimator}
\label{sec:nocs2pose}
\vspace{-.2em}

Conventional Structure-from-Motion (SfM) solvers, such as COLMAP~\cite{schonberger2016structure}, rely on feature matching for pose estimation. However, in scenarios with sparse views, there may be little to no overlap between input views. The lack of sufficient visual correspondence cues often renders the solvers unreliable and prone to failure. Consequently, instead of relying on local correspondences, we leverage the rich semantic priors embedded in 2D diffusion models for pose estimation.

One of the primary challenges is to enable 2D diffusion models to output camera poses. While camera poses can be represented in various scalar formats (e.g., 6-dimensional vector, four-by-four matrix, etc.), they are not native representations for a 2D diffusion model to generate. Inspired by recent works demonstrating that 2D diffusion models can be used to predict normal maps~\cite{long2023wonder3d}—a domain different from natural images—we propose using a surrogate representation: the Normalized Object Coordinate Space (NOCS)~\cite{wang2019normalized}. We finetune Stable Diffusion to predict NOCS maps for each input view.

As depicted in Fig.~\ref{fig:inputoutput}(b), a NOCS frame is determined for each set of input sparse view images and the underlying 3D object. Specifically, the 3D shape is normalized into a unit cube, i.e., ${x,y,z} \in [0,1]$. The shape's upward axis aligns with the dataset's inherent upward axis of the 3D object, typically the gravity axis. Predicting the object's forward-facing direction may be ambiguous, so we rotate the 3D shape in the NOCS frame to align its forward direction (zero azimuth) with that of the first input view, thus unambiguously establishing the NOCS frame. For each input view, we then render a NOCS map, where each 2D pixel (\textit{r,g,b}) represents the corresponding 3D point's position (\textit{x,y,z}) in the defined NOCS frame, as shown in Fig.~\ref{fig:inputoutput}(c). These NOCS maps align with the operational domain of 2D diffusion models, similar to the normal maps in previous work~\cite{long2023wonder3d}.

To facilitate interactions between NOCS maps from different views and generate more 3D-consistent NOCS maps, we tile all NOCS maps into a $3\times 2$ grid image as the input condition (see~\cref{sec:tiling}), following the same tiling order and the empty padding convention. We finetune Stable Diffusion to generate these multi-view tiled NOCS maps, so the 2D diffusion model can attend to both the input sparse views and their NOCS maps during the diffusion process. 

After generating the NOCS maps for the input sparse views, we employ a traditional Perspective-n-Point (PnP) solver~\cite{opencvpnp} to compute the poses $\{\boldsymbol{\xi}_i^{\text{pnp}}\}$ from the NOCS frame to the camera frames of each input view by minimizing the reprojection error:
\vspace{-.9em}
\begin{equation} \label{eq_pnp}
\boldsymbol{\xi}_i^{\text{pnp}} =
\arg\min_{\boldsymbol{\xi}_i \in \text{SE(3)}} \sum_{j=1}^{m_i} \| \mathbf{p}_{i,j} - \text{proj}(\mathbf{q}_{i,j}, \boldsymbol{\xi}_i) \|_2^2,
\vspace{-.5em}
\end{equation}
where $\mathbf{p}_{i,j}$ represents the $j^{\text{th}}$ pixel's location in the $i^{\text{th}}$ NOCS map; $\mathbf{q}_{i,j}$ is the corresponding 3D point location in the NOCS frame; $m_i$ is the number of pixels for the $i^{\text{th}}$ view, and $\text{proj}$ is the perspective projection operation. Note that the PnP algorithm assumes known camera intrinsics and optimizes only for the camera extrinsics. A RANSAC scheme is applied during the PnP computation for outlier removal, enhancing the robustness of the pose prediction to boundary noises and errors from the 2D diffusion model. As all NOCS maps share a common NOCS frame, we can thus determine the relative camera poses between views $i$ and $i^{\prime}$ through $\boldsymbol{\xi}_i^{-1}\boldsymbol{\xi}_{i^{\prime}}$. 

\vspace{-.5em}
\subsection{Multi-View Prediction for 3D Reconstruction}
\label{sec:recon}
\vspace{-.2em}

We follow the paradigm of recent single-image-to-3D methods~\cite{liu2023one2345++, li2023instant3d} by initially generating multi-view images and subsequently using a feed-forward 3D reconstruction module to convert these images into a 3D representation. It is noteworthy that the input sparse views might not encompass the entire 3D objects, nor provide adequate information for 3D reconstruction. Therefore, we propose to predict multi-view images at uniformly distributed camera poses first, and then use these predicted images for 3D reconstruction.

Unlike traditional novel view synthesis~\cite{liu2023zero}, our approach employs a fixed camera configuration for target multi-views. As depicted in~\cref{fig:inputoutput}, our target multi-view images consist of six views with alternating $20^{\circ}$ and $-10^{\circ}$ elevations, and $60^{\circ}$-spaced azimuths relative to the first input view. Although the elevation angles are set absolutely, the azimuth angles are relative to the azimuth of the first input sparse view to resolve the ambiguity in face-forwarding directions. Furthermore, We maintain consistent camera intrinsics across target views, independent of input views. These strategies mitigate challenges in predicting camera intrinsics and elevation during the 3D reconstruction process. Existing methods hindered by this issue may be sensitive to intrinsic variations and often depend on predicting~\cite{liu2023one2345} or requiring user-specified~\cite{liu2023syncdreamer,tang2023dreamgaussian,tang2024lgm} input image elevations.

Similar to NOCS map prediction, we tile all six views into a $3\times 2$ grid image and finetune Stable Diffusion to generate this tiled image. The 2D diffusion model, conditioned on the input sparse views, aims to incorporate all information from input views, deduce the underlying 3D objects, and predict the multi-view images at the predetermined camera poses. Although the predicted poses of input sparse views (\cref{sec:nocs2pose}) are not directly employed in the 3D reconstruction, the joint training of NOCS prediction and multi-view prediction branches implicitly complement each other and boost the performance of both tasks.

Upon generating the multi-view images at known camera poses, we utilize the multi-view to 3D reconstruction module proposed in~\cite{liu2023one2345++} to lift these images to 3D. The reconstruction module adopts a two-stage coarse-to-fine approach, which involves initially extracting the 2D features of the generated multi-view images, aggregating them with the known camera poses, and constructing a 3D cost volume. This 3D cost volume acts as the condition for the 3D diffusion networks. In the coarse stage, a low-resolution $64^3$ 3D occupancy volume is produced. This is subsequently refined to yield a high-resolution $128^3$ SDF (Signed Distance Field) volume with colors. Finally, a textured mesh is derived from the SDF volume employing the marching cubes algorithm.

\vspace{-.5em}
\subsection{Pose Refinement with Reconstructed 3D Model}
\label{sec:pose_refine}
\vspace{-.2em}

In Section~\ref{sec:nocs2pose}, we finetune diffusion models for NOCS map prediction and camera pose estimation. However, due to the hallucinatory and stochastic nature of diffusion models, unavoidable errors may exist. The generated 3D mesh $\mathcal{M}$, though not perfect, provides a multi-view consistent and explicit 3D structure. We can further refine the coarse poses predicted from the NOCS maps by leveraging the reconstructed 3D shape.

\noindent\textbf{Pose Refinement via Differentiable Rendering.}
Starting with initial poses $\{\boldsymbol{\xi}_i^{\text{pnp}}\}$ extracted from the predicted NOCS maps, we refine them through differentiable rendering \cite{Laine2020diffrast}. Specifically, we render the generated mesh $\mathcal{M}$ at optimizing camera poses $\boldsymbol{\xi}_i$. We minimize the rendering loss between the rendered image $\mathbf{I}^{r}_i = \mathcal{R}(\mathcal{M}, \boldsymbol{\xi}_i)$ and the input image $\mathbf{I}_i$ to obtain the optimally fitted camera pose $\boldsymbol{\xi}^*_i$. The optimization process can be formulated as: 
\vspace{-.5em}
\begin{equation}
\label{eq_pose}
\boldsymbol{\xi}^*_i = \arg\min_{\boldsymbol{\xi}_i \in \text{SE(3)}} \, (\lambda \cdot \mathcal{L}_{\text{mask}}(\mathbf{I}^{r}_i, \mathbf{I}_i) + \mu \cdot \mathcal{L}_{\text{rgb}}(\mathbf{I}^{r}_i, \mathbf{I}_i)),
\end{equation}

where $\mathcal{L}_{\text{mask}}$ and $\mathcal{L}_{\text{rgb}}$ are the cross-entropy and MSE losses computed for the foreground masks and the RGB values, respectively, and $\lambda$ and $\mu$ are two weighting coefficients. The refinement process is lightweight and can be completed in just one second, given the generated mesh.

\noindent\textbf{Mixture of Experts (MoE).} The NOCS pose predictions are inherently stochastic and may not produce an accurate pose in a single pass. For instance, with objects possessing certain symmetries, the diffusion model may predict only one of the possible symmetric poses. We employ a Mixture of Experts (MoE) strategy to further refine the pose, which is simple but effective. Specifically, we generate multiple NOCS maps for each input view using different seeds. We then select the pose that minimizes the rendering loss based on the refinement results with the generated 3D mesh. This technique effectively reduces pose estimation error, as quantitatively validated by the ablation study in the \suppmat.

\vspace{-2em}
\section{Experiments}
\label{sec:exp}
\vspace{-0.6em}
\subsection{Evaluation Settings}
\vspace{-0.4em}

\begin{table}[t]

    \scriptsize
  \caption{ 
  \textbf{Evaluation Results for Pose Estimation.} We compare our method with RelPose++~\cite{lin2023relpose++}, FORGE~\cite{jiang2022few}, and iFusion~\cite{wu2023ifusion} on three unseen datasets: OmniObject3D~\cite{wu2023omniobject3d}, GSO~\cite{downs2022google}, and ABO~\cite{collins2022abo}. $500$ objects are sampled for each dataset.
  }
  \vspace{-1em}
  \centering
  \begin{tabular}{llccccc}
    \toprule
    Dataset & Method & Rot. Err$\downarrow$ & Acc.@15$^{\circ}\uparrow$ & Acc.@30$^{\circ}\uparrow$ & Trans. Err$\downarrow$ & Time$\downarrow$ \\
    
    \midrule
    
    \multirow{7}{*}{GSO~\cite{downs2022google}}
        & RelPose++ & 103.24 & 0.011& 0.033 & 4.84 &  3.6s \\
        & FORGE& 111.40 & 0.004 & 0.020 & 4.21 &  440s \\
        & iFusion ($n_{init}$=1) & 95.15 & 0.208 & 0.258 & 3.65 &  64s \\ 
        & iFusion ($n_{init}$=4) & 8.61 & 0.651 & 0.759 & 0.49 &  256s \\ 
        & Ours (w/o refine) & 13.02 & 0.537 & 0.616 & 0.58 & 10s  \\
        & Ours ($n_{init}$=1) & 9.87 & 0.563 & 0.617 & 0.42 &  27s \\
        & Ours ($n_{init}$=4) & \textbf{5.28} & \textbf{0.750} & \textbf{0.787} & \textbf{0.23} & 57s \\

    \midrule

    \multirow{7}{*}{OO3D~\cite{wu2023omniobject3d}}
        & RelPose++ & 105.05 & 0.008 & 0.046 & 7.38 & 3.6s \\
        & FORGE & 99.27 & 0.014 & 0.063 & 7.27 & 440s\\
        & iFusion ($n_{init}$=1) & 91.15 & 0.166 & 0.271 & 4.77 & 64s \\
        & iFusion ($n_{init}$=4) & 15.08 & 0.498 & 0.721 & 1.12 & 256s \\
        & Ours (w/o refine) & 14.75 & 0.508 & 0.725 & 0.90 & 10s \\
        & Ours ($n_{init}$=1) & 13.40 & 0.544 & 0.730 & 0.89 & 27s \\
        & Ours ($n_{init}$=4) & \textbf{10.07} & \textbf{0.668} & \textbf{0.849} & \textbf{0.63} & 57s \\

    \midrule
    
    \multirow{7}{*}{ABO~\cite{collins2022abo}}
        & RelPose++ & 103.14 & 0.017& 0.039 & 5.01 & 3.6s\\
        & FORGE & 110.64 & 0.005 & 0.023 & 4.18 & 440s \\
        & iFusion ($n_{init}$=1) & 96.65  & 0.186 & 0.219 & 3.88 & 64s \\
        & iFusion ($n_{init}$=4) & 8.55 & 0.578 & 0.631 & 0.68 & 256s \\
        & Ours (w/o refine) & 10.87 & 0.554 & 0.597 & 0.49 & 10s \\
        & Ours ($n_{init}$=1) & 9.30 & 0.565 & 0.600 & 0.43 & 27s \\
        & Ours ($n_{init}$=4) & \textbf{5.80} & \textbf{0.675} & \textbf{0.701} & \textbf{0.27} & 57s \\
    \bottomrule
  \end{tabular}
  \label{tab:pose_comp}
  \vspace{-1em}
\end{table}

\noindent\textbf{Training Datasets and Details.} 
We train our models on a curated subset of 100k shapes from the Objaverse dataset~\cite{deitke2023objaverse}. Considering the variable quality of the original Objaverse dataset, we opted to filter out higher-quality data by initially manually annotating 8,000 3D objects based on overall geometry quality and texture preferences. Subsequently, we train MLP models for quality rating classification and texture score regression, utilizing their multimodal features~\cite{liu2024openshape}. Based on the predictions of these models, we select shapes that are rated as high-quality and have top texture scores. Further details about the data filtering process are included in the \suppmat.

For each 3D shape, we render 10 sets of images using BlenderProc~\cite{denninger2019blenderproc}. Each set comprises 6 input images, 6 output multi-view images, and 6 NOCS maps. To mimic real-world conditions and ensure model robustness, we randomly sample camera intrinsics and extrinsics, as well as environment maps for the input images. For the output multi-view images, their intrinsics remain constant, while the extrinsics are derived from a fixed delta pose and the azimuth of the input images. Each set of input and output images shares the same environment map. During training, we randomly selected between 1 to 6 views as sparse input views, with the first view of each set always being included. We train the model utilizing 8 A100 GPUs for approximately 3 days.

\begin{figure}[t]
    \centering
    \includegraphics[width=1\linewidth]{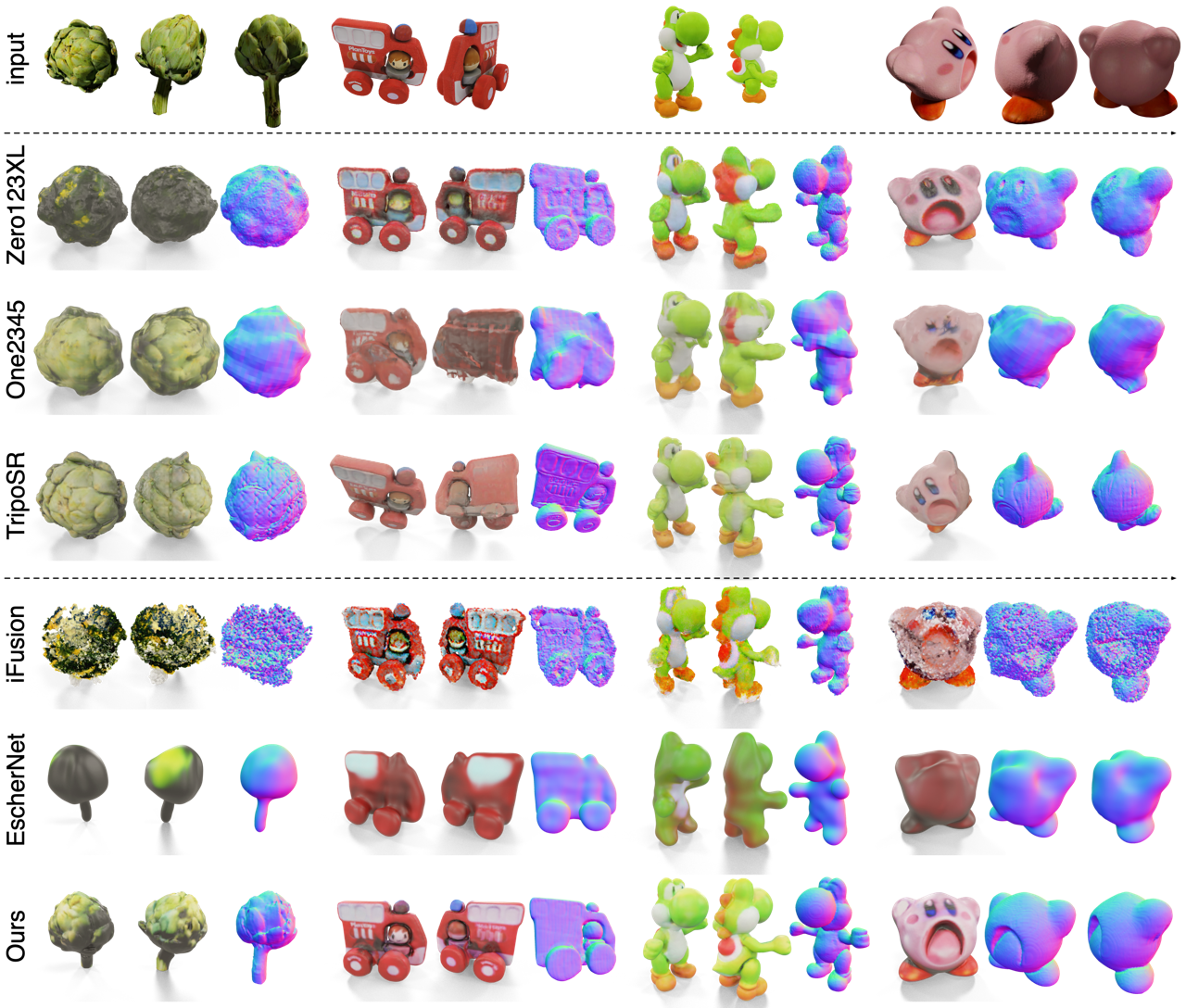}
    \vspace{-2em}
    \caption{\textbf{Qualitative Results on 3D Reconstruction.} Zero123XL~\cite{deitke2023objaversexl}, One2345~\cite{liu2023one2345}, and TripoSR~\cite{tochilkin2024triposr} are single-image-to-3D methods, each utilizing only the first input image. iFusion~\cite{wu2023ifusion}, EscherNet~\cite{kong2024eschernet}, and our approach take all input images (the first row). Textured meshes and mesh normal renderings are shown. Shapes come from the OmniObject3D~\cite{wu2023omniobject3d} and GSO~\cite{downs2022google} datasets.
    }
    \vspace{-2em}
    \label{fig:comp_3d}
\end{figure}

\noindent\textbf{Baselines.} 
For 3D reconstruction, we compare our method with both state-of-the-art single-image-to-3D and sparse-view-to-3D baselines. Single-view-to-3D methods we evaluate include optimization-based approaches, such as Zero123 XL~\cite{deitke2023objaversexl}, SyncDreamer~\cite{liu2023syncdreamer}, and DreamGaussian~\cite{tang2023dreamgaussian}, as well as feed-forward methods like One-2-3-45~\cite{liu2023one2345} and Shap-E~\cite{jun2023shap}. For sparse-view methods, we consider two recent open-source approaches as baselines: iFusion~\cite{wu2023ifusion} and EscherNet~\cite{kong2024eschernet}. We utilize ThreeStudio~\cite{threestudio2023}'s implementation for Zero123 XL and the official implementations for the other baselines. Specifically, for iFusion, we use their official reconstruction pipeline integrated with Zero123 XL.

For sparse-view pose estimation, we compare our method with state-of-the-art approaches including RelPose++~\cite{lin2023relpose++}, FORGE~\cite{jiang2022few}, and iFusion~\cite{wu2023ifusion}. The latter two are optimization-based while \cite{lin2023relpose++} is a feed-forward method.

\noindent\textbf{Evaluation Datasets.} 
For 3D reconstruction, we evaluate the methods on the entire GSO~\cite{downs2022google} dataset, which comprises 1,030 3D shapes; none of these shapes were seen during our training. For each 3D shape, we randomly render six views as input images. For single-image-to-3D methods, a fixed-view image is taken as input following~\cite{liu2023one2345}. We carefully align the predictions with the ground truth meshes before calculating the metrics. Please refer to the \suppmat for detailed information on shape alignment and the evaluation metrics.

For pose estimation, we evaluate the approaches on three datasets: OmniObject3D~\cite{wu2023omniobject3d} and GSO~\cite{downs2022google}, both captured from real scans, and ABO~\cite{collins2022abo}, a synthetic dataset created by artists. For each dataset, we randomly choose 500 objects and render five random sparse views per shape. We follow iFusion~\cite{wu2023ifusion} to report the rotation accuracy and the median error in rotation and translation across all image pairs. More details are provided in the \suppmat.

\begin{table}[t]
\scriptsize
\setlength{\tabcolsep}{2pt}
  \centering
  \caption{\textbf{Quantitative Comparison on 3D Reconstruction.} Evaluated on the complete GSO~\cite{downs2022google} dataset, which contains 1,030 3D objects. Five single-image-to-3D methods~\cite{deitke2023objaversexl,jun2023shap,liu2023one2345,liu2023syncdreamer,tang2023dreamgaussian} and two sparse-view methods~\cite{wu2023ifusion,kong2024eschernet} are compared.} 
  \vspace{-1em}
  \begin{tabular}{c|c|ccccc}
    \toprule
    $n_{input}$ & Method & F-Score (\%)$\uparrow$ & CLIP-Sim$\uparrow$ & PSNR$\uparrow$ & LPIPS$\downarrow$ & Time$\downarrow$ \\
    \midrule
    
    \multirow{5}{*}{1}
    
    & Zero123 XL~\cite{deitke2023objaversexl} &  91.6    &   73.1  &  18.16 & 0.136 &  20min\\
    & Shap-E~\cite{jun2023shap} & 91.8  & 73.1  &  18.96   & 0.140 & 27s \\
    & One-2-3-45~\cite{liu2023one2345} &   90.4    & 70.8      &    19.07  & 0.133 & 45s \\
    & SyncDreamer~\cite{liu2023syncdreamer} & 84.8  &    68.9   &  16.86  &  0.145 & 6min \\
    & DreamGaussian~\cite{tang2023dreamgaussian} & 81.0 & 68.4 & 17.88 &  0.147 & 2min \\
    & Ours &  \textbf{95.7}  & \textbf{78.2} & \textbf{19.87} & \textbf{0.124} & \textbf{16s} \\
    \midrule
    
    \multirow{3}{*}{6} 
    
    & iFusion~\cite{wu2023ifusion} & 88.5  & 66.7 &  16.2 & 0.151 & 28min \\
    & EscherNet~\cite{kong2024eschernet} & 94.8 & 65.9 & 16.6 & 0.139 & 9min \\
    
    & Ours  & \textbf{96.9} & \textbf{78.1} &    \textbf{19.3} & \textbf{0.123} & \textbf{16s} \\
    \bottomrule
  \end{tabular}
  \vspace{-1.2em}
  \label{tab:3d_comp}
\end{table}

\begin{figure}[t]
    \centering
    \includegraphics[width=1\linewidth]{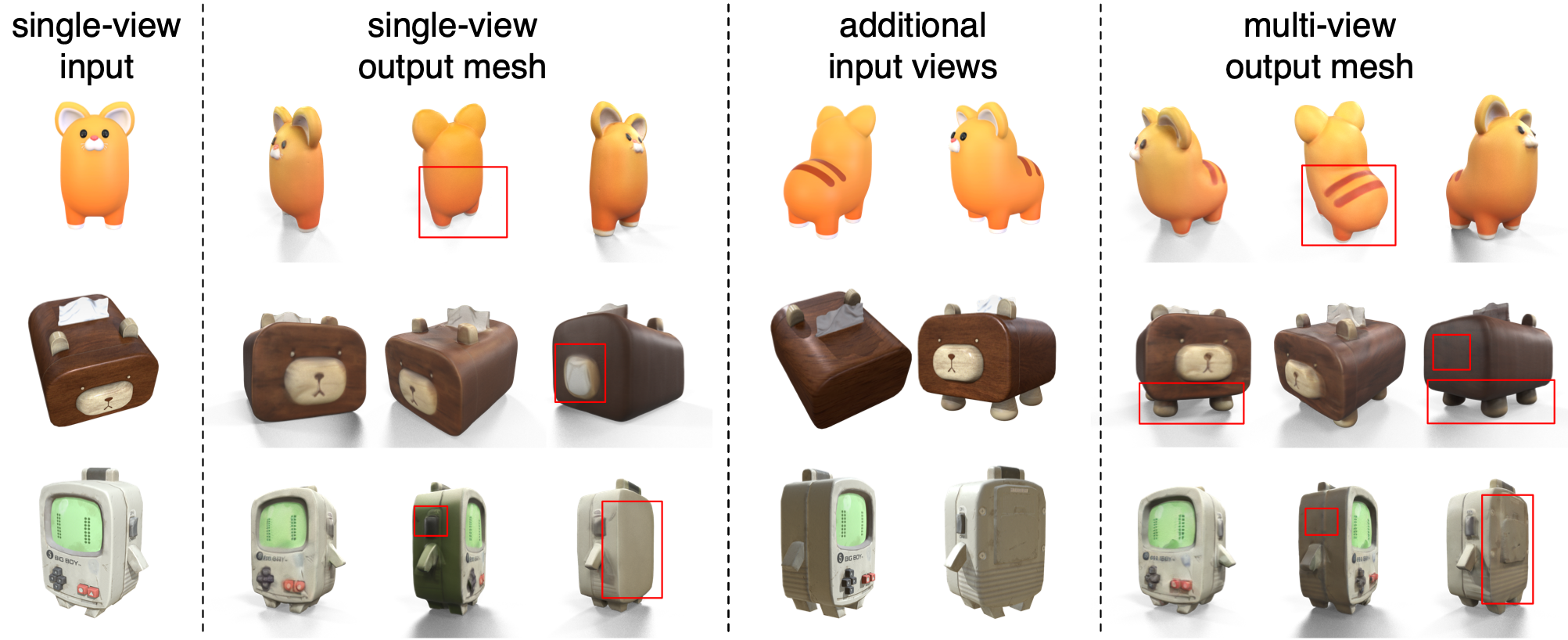}
    \vspace{-1.9em}
    \caption{\textbf{Single-View vs. Sparse-View for 3D Reconstruction.} We compare the results of our method when using single-view and sparse-view inputs. }
    \vspace{-1.5em}
    \label{fig:single_vs_multi}
\end{figure}

\begin{table}[t]
\scriptsize
  \caption{\textbf{Ablation Study on the Number of Input Views.} Evaluated on the GSO dataset~\cite{downs2022google}.}
  \centering
  \vspace{-1em}
  \begin{tabular}{c|ccc|ccc}
    \toprule
    
    $n$ views & Rot. Err$\downarrow$ & Acc.@5$^{\circ}\uparrow$ & Trans. Err$\downarrow$ & F-Score (\%)$\uparrow$ & CLIP-Sim$\uparrow$ & PSNR$\uparrow$ \\
    
    \midrule
    
    1 & -- & -- & -- & 89.1 & 74.9 &  17.7  \\

    2 & 8.56 & 0.32 & 0.42 & 93.3 & 76.5 &  18.3  \\

    4 & 6.03 & 0.43 & 0.28 & 96.0 & 77.6 &  19.0  \\

    6 & \textbf{5.28} & \textbf{0.48} & \textbf{0.25} & \textbf{96.9} & \textbf{78.1} &    \textbf{19.3}    \\
    \bottomrule
  \end{tabular}
  \vspace{-1em}
  \label{tab:ab_nview}
\end{table}

\begin{figure}[h]
    \centering
    \includegraphics[width=1\linewidth]{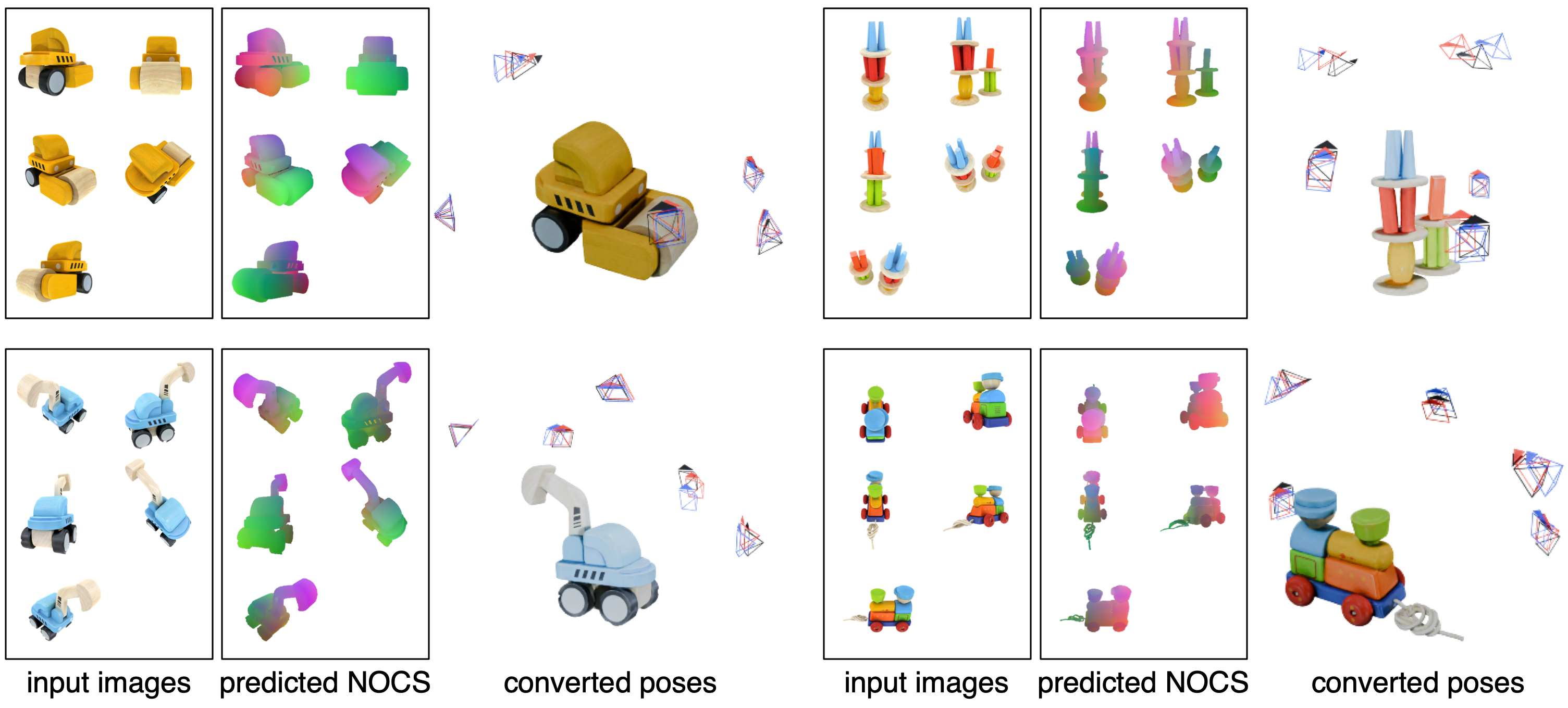}
    \vspace{-2em}
    \caption{\textbf{Ablation Study on Pose Refinement.} We showcase the input images, predicted NOCS maps, and converted poses. The ground truth poses are in black, while the predicted poses before and after refinement are in blue and red, respectively.}
    \label{fig:pose_refine}
    \vspace{-1.5em}
\end{figure}

\begin{table}[t]
\scriptsize
  \caption{\textbf{Effect of Joint Training.} Evaluated on 500 objects from GSO~\cite{downs2022google}.}
  \centering
  \vspace{-.5em}
  \begin{tabular}{c|cccc|ccc}
    \toprule
    Method & Rot. Err$\downarrow$ & Acc.@15$^{\circ}\uparrow$ & Acc.@30$^{\circ}\uparrow$ & Trans. Err$\downarrow$ & F-Score (\%)$\uparrow$ & CLIP-Sim$\uparrow$ & PSNR$\uparrow$ \\
    \midrule
    Separate & 9.87  & 0.563   &  0.617 &  0.42 &  96.81  & 78.36 & 19.03  \\
    Joint & \textbf{8.57} & \textbf{0.601} & \textbf{0.677} & \textbf{0.37} & \textbf{97.12} & \textbf{78.90} & \textbf{19.42}  \\
    \bottomrule
  \end{tabular}
  \vspace{-1em}
  \label{tab:ab_mixed}
\end{table}

\vspace{-.8em}
\subsection{Experiment Results}
\vspace{-.3em}
\noindent\textbf{Pose Prediction.} 
We report the pose estimation results in \cref{tab:pose_comp}, where it is evident that \ours outperforms all baseline methods by a significant margin. It is worth noting that RelPose++~\cite{lin2023relpose++} and FORGE~\cite{jiang2022few} struggle to yield satisfactory results for our open-world evaluation images. iFusion, an optimization-based approach, is prone to becoming trapped in local minima. With only one initial pose ($n_{init}=1$), it also fails to produce adequate results. In contrast, our method leverages priors from 2D diffusion models and can generate acceptable results in a single forward pass. Even without any additional refinement (w/o refine), our method can already produce results similar to iFusion with four initial poses ($n_{init}=4$), while being far more efficient, requiring just 1/25 of the runtime. With the integration of further refinement through a mixture of experts, our method achieves even better performance.

\noindent\textbf{3D Reconstruction.} 
We present the qualitative results in \cref{fig:comp_3d}. With only a single-view input, single-image-to-3D methods fail to produce meshes that faithfully match the entire structure and details of the ground truth mesh. For instance, most single-view baseline methods are unable to reconstruct the stems of the artichoke, the back of the firetruck, the red saddle on Yoshi, and the two separate legs of Kirby standing on the ground. In contrast, sparse-view methods yield results that are much closer to the ground truth by incorporating information from multiple sparse views. Compared to iFusion, EscherNet, our method generates meshes with higher-quality geometry and textures that more accurately match the input sparse views. We report the quantitative results in \cref{tab:3d_comp}, where our method significantly outperforms both single-view-to-3D and sparse-view approaches in terms of both 2D and 3D metrics. Moreover, our method exhibits superior efficiency, being much faster than the baseline methods.

\vspace{-.8em}
\subsection{Analysis} 
\vspace{-0.3em}

\noindent\textbf{Single View vs. Sparse Views.} In \cref{fig:single_vs_multi}, we present the results obtained by our method when provided with single-view and sparse-view inputs. With a single-view input, our method can still generate reasonable results, yet it may not accurately capture the structures and details of the regions that are not visible. Our method demonstrates the capability to effectively integrate information from all sparse-view inputs provided.

\noindent\textbf{Number of Views.} In \cref{tab:ab_nview}, we quantitatively showcase the impact of the number of views on both 3D reconstruction and pose estimation. We observe that incorporating more input views enables the 2D diffusion network to better grasp their spatial relationships and underlying 3D objects, boosting both tasks.

\noindent\textbf{Pose Refinement.} While the predicted NOCS maps can be directly converted into camera poses, we have found that these poses can be further refined through alignment with the generated 3D meshes. \cref{fig:pose_refine} showcases the predicted poses before and after refinement. Although both are generally very close to the ground truth poses, refinement can further reduce the error.

\noindent\textbf{Number of Experts.} We employ a mixture-of-experts strategy to address the ambiguity issues related to NOCS prediction for symmetric objects. By using this strategy and increasing the number of experts, there is a substantial increase in pose estimation accuracy. Please refer to the \suppmat for more details and quantitative ablation studies.

\noindent\textbf{Joint Training.} We finetune 2D diffusion models to jointly predict NOCS maps and multi-view images from sparse, unposed views by leveraging a domain switcher. As shown in \cref{tab:ab_mixed}, this joint training strategy enables the two branches to implicitly interact and complement each other, enhancing the interpretation of both the input sparse views and the intrinsic properties of the 3D objects, which in turn improves the performance of each task.

\vspace{-.6em}
\section{Conclusion}
\vspace{-.4em}
\label{sec:conclusion}

We present \ours, a novel method for 3D reconstruction and pose estimation using unposed sparse-view images. Our method leverages rich priors embedded in 2D diffusion models and exhibits strong open-world generalizability. Without the need for per-shape optimization, it can deliver high-quality textured meshes, along with accurate camera poses, in approximately 20 seconds.\\

\textbf{Acknowledgements: } We thank Chong Zeng, Xinyue Wei for the discussion and help with data processing, and Peng Wang for providing the evaluation set. We also extend our thanks to all annotators for their meticulous annotations.

%
%
\bibliographystyle{splncs04}
\bibliography{main}

\newpage
\appendix
\vspace{-2em}
\begin{figure}[H]
    \centering
     \includegraphics[width=\linewidth]{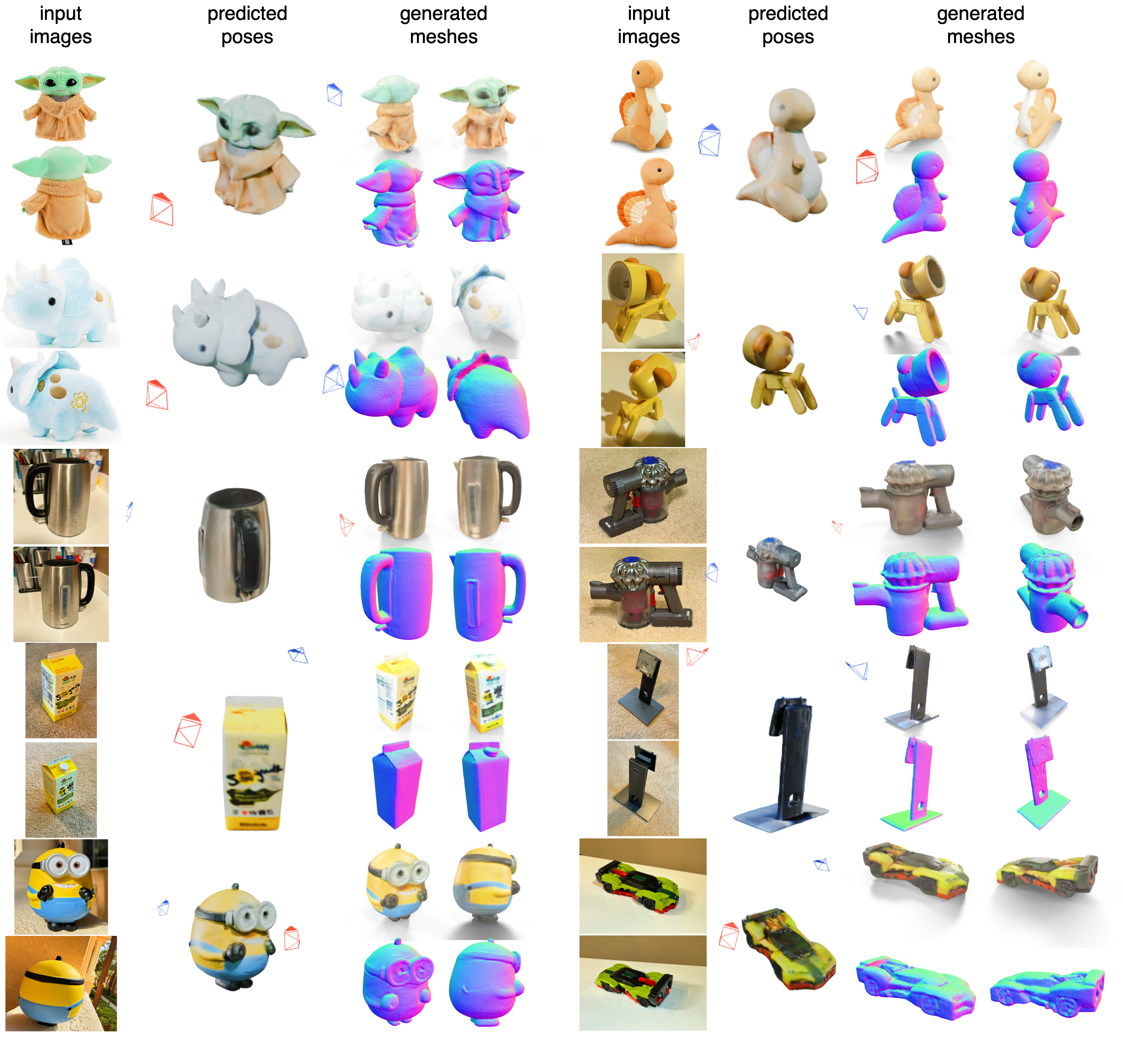}
    \caption{\textbf{Real-World Examples}: The input images are either sourced from amazon.com or captured using an iPhone.} 
    \label{fig:real-world}
\end{figure}

\section{Additional Real-World Examples}

In \cref{fig:real-world}, we demonstrate that \ours can be applied to real-world sparse-view images without camera poses. This includes images captured by users with consumer devices (e.g., with an iPhone) or e-commerce product images (e.g., from amazon.com). \ours is capable of achieving commendable results in both pose estimation and 3D reconstruction.

\section{Ablation Studies on Number of Experts} 
\label{sec:moe}

\begin{table}[t]

  \caption{\textbf{Ablation Study on the Number of Experts for Pose Estimation.} Evaluated on the OmniObject3D~\cite{wu2023omniobject3d} and GSO~\cite{downs2022google} datasets.}
  \centering
  \begin{tabular}{c|ccc|ccc}
    \toprule
    \multirow{2}{*}{\makecell{$n_{init}$ \\experts}}
    & \multicolumn{3}{c}{OmniObject3D~\cite{wu2023omniobject3d}}   &  \multicolumn{3}{c}{Google Scanned Objects~\cite{downs2022google}} \\
    \cmidrule{2-7}
     & Rot. Err $\downarrow$ & Acc.@15$^{\circ}\uparrow$ & Trans. Err$\downarrow$ & Rot. Err $\downarrow$ & Acc.@15$^{\circ}\uparrow$ & Trans. Err$\downarrow$ \\
    
    \midrule
    
    1 & 15.23 & 0.495 & 1.04 & 9.83 & 0.562 & 0.43 \\

    2 & 12.05 & 0.585 & 0.76 & 6.03 & 0.687 & 0.28 \\
    
    4 & 10.46 & 0.647 & 0.68 & 4.71 & 0.805 & 0.21 \\
    
    8 & \textbf{9.46} & \textbf{0.690} & \textbf{0.60} & \textbf{4.34} & \textbf{0.853} & \textbf{0.20} \\
    \bottomrule
  \end{tabular}
  \label{tab:ab_moe}
\end{table}

Due to the inherently stochastic nature of diffusion models, our multi-view diffusion model may sometimes fail to accurately understand the spatial relationship between input images of objects and estimate their relative poses in a single diffusion pass, especially with objects that have some symmetry. We found that employing a Mixture of Experts (MoE) strategy effectively mitigates this issue. Specifically, we run the diffusion models $n_{init}$ times with different random seeds to generate multiple
sets of NOCS maps for pose prediction, selecting the optimal one based on the minimum rendering loss from the pose refinement stage. As shown in~\cref{tab:ab_moe}, increasing the number of experts ($n_{init}$) from 1 to 8 led to a significant improvement in the accuracy of relative pose predictions across both the OmniObject3D~\cite{wu2023omniobject3d} and GSO~\cite{downs2022google} datasets. This demonstrates that the MoE strategy is simple yet effective in improving the robustness of our pose prediction approach.

\section{Robustness to Varying Camera Intrinsics}
\begin{figure}[t]
    \centering
     \includegraphics[width=\linewidth]{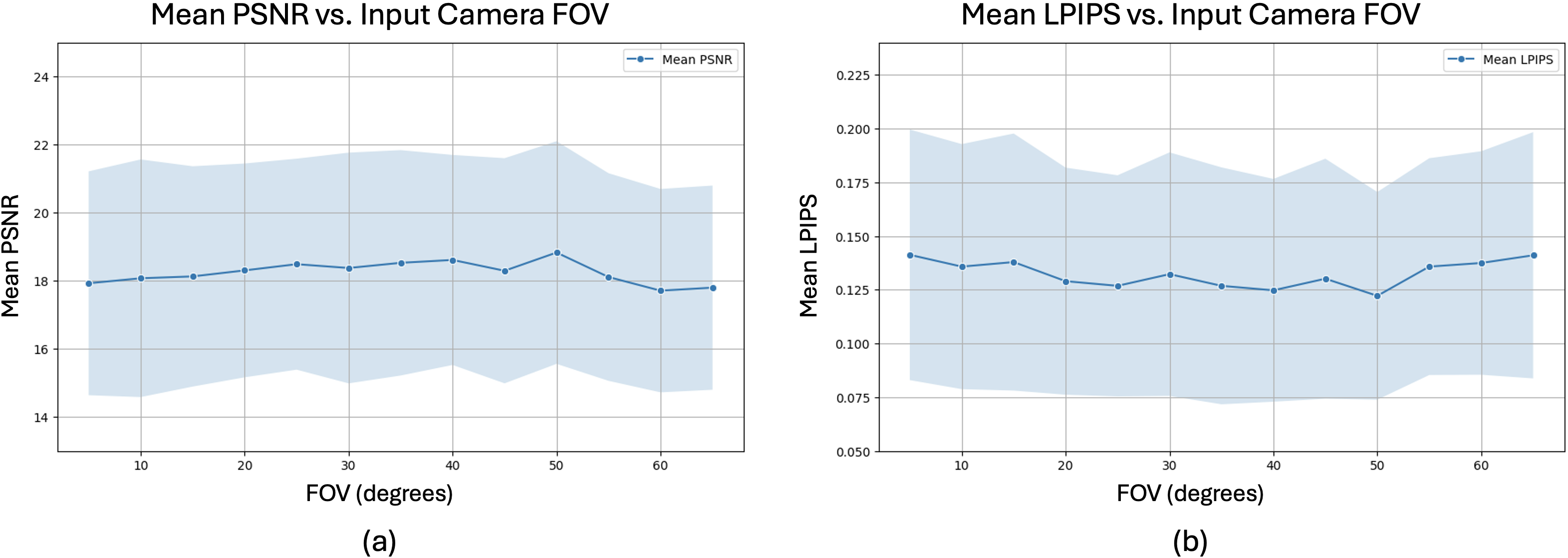}
    \caption{Our model achieves consistently high PSNR and low LPIPS across different input-view FOVs, with no significant performance degradation due to focal length variations.} 
    \label{fig:intrinsics}
\end{figure}

Our multi-view diffusion model demonstrates robust performance across varying input image camera intrinsics. During its training, we randomize both the focal length and optical center of input images. The input image field of view (FOV) follows a normal distribution $\mathcal{N}(36^\circ, 9^\circ)$, centered at 36 degrees. The optical center also follows a normal distribution centered at the image center.
As shown in ~\cref{fig:intrinsics},
we tested the model's performance across input FOVs ranging from 5 to 65 degrees, covering common photographic focal lengths. Using 20 different objects, we calculated the average PSNR and LPIPS for predictions at various FOVs. Our model demonstrated consistently high performance across the tested range. This showcases its robustness to intrinsic variations in input images.

\section{Sparse-View Reconstruction using the Estimated Poses}

\begin{figure}[t]
    \centering
     \includegraphics[width=\linewidth]{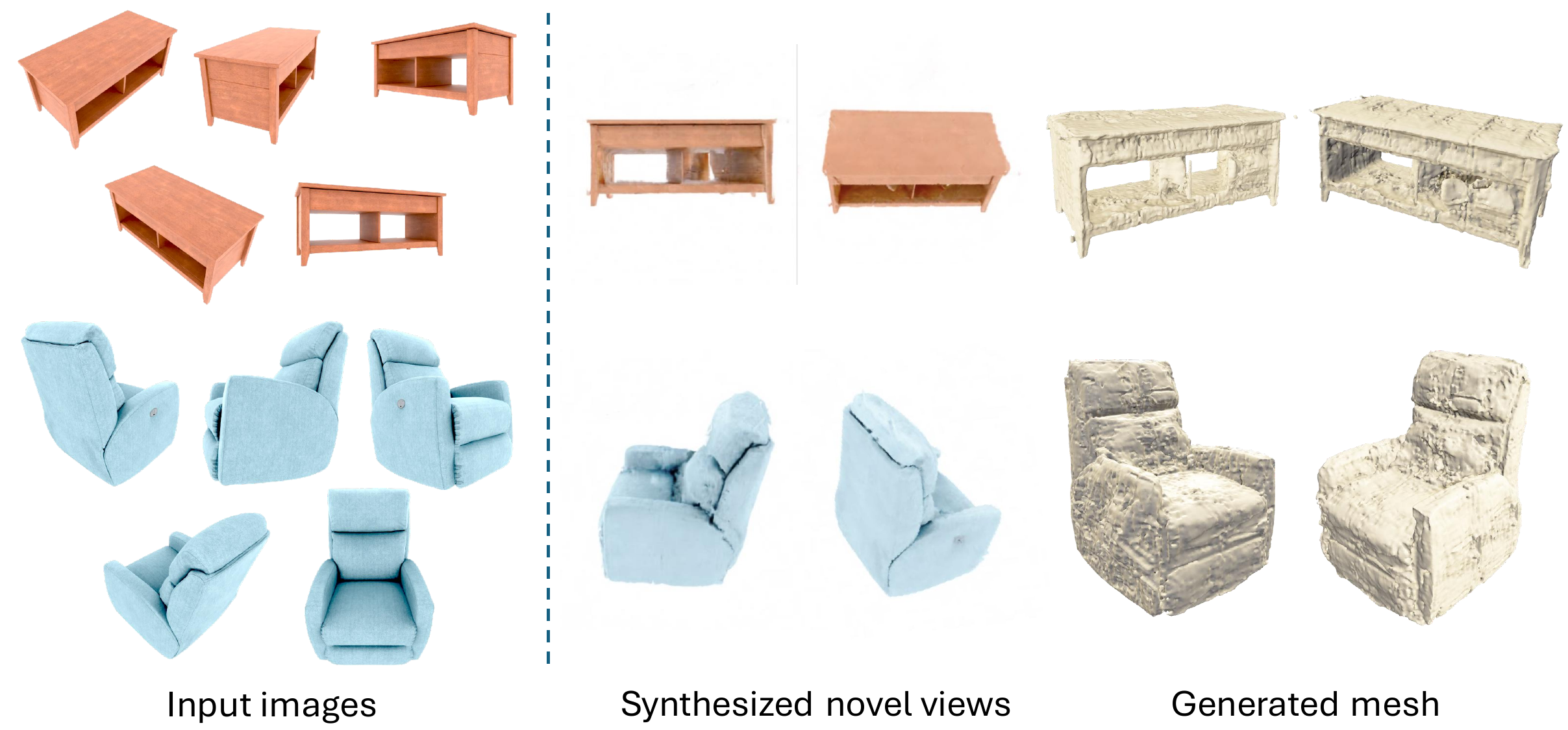}
    \caption{The camera poses predicted by our method can be utilized in ZeroRF~\cite{shi2023zerorf}, which is an optimization-based sparse-view reconstruction method requiring camera poses as input. The input images are sourced from the ABO dataset~\cite{collins2022abo}.} 
    \label{fig:zerorf}
\end{figure}

Our estimated poses can benefit numerous downstream applications, including many existing sparse-view 3D reconstruction approaches that require camera poses. Here, we demonstrate how our estimated poses can be utilized with ZeroRF~\cite{shi2023zerorf}, a sparse-view 3D reconstruction method. ZeroRF is an optimization-based method that does not rely on pretrained priors and requires camera poses as input. As depicted in~\cref{fig:zerorf}, by using only five images along with the corresponding predicted poses as input, ZeroRF is capable of generating a NeRF that synthesizes reasonable novel views. The resulting mesh also shows commendable global geometry, considering the challenging nature of the task.

\section{Evaluation Details}
\label{sec:metrics}

\subsection{3D Reconstruction}
To account for the scale and pose ambiguity of the generated mesh, we align the predicted mesh with the ground truth mesh prior to metric calculation. During the alignment process, we sample 12 rotations ($30^\circ$ apart) as initial positions and 10 scales from 0.6 to 1.4, which is dense enough in practice. We enumerate the combinations of these rotations and scales for initialization and subsequently refine the alignment with the Iterative Closest Point (ICP) algorithm. We select the alignment that yields the highest inlier ratio. Both the ground truth and predicted meshes are then scaled to fit within a unit bounding box.

We adopt the evaluation metrics from~\cite{liu2023one2345++} to assess the reconstruction quality from two perspectives: (1) geometric quality and (2) texture quality. For geometric quality, we apply the F-score to quantify the discrepancy between the reconstructed and ground truth meshes, setting the F-score threshold at 0.05. To evaluate texture quality, we compute the CLIP-Similarity, PSNR, and LPIPS between images rendered from the reconstructed mesh and those of the ground truth. The meshes undergo rendering from 24 distinct viewpoints, encompassing a full 360-degree view around the object. The rendered images have a resolution of 512 $\times$ 512 pixels.

\subsection{Pose Estimation}
To evaluate pose estimation, we render five sparse views for each shape and assess the relative poses between all ten pairs of views. We convert the predicted poses to the OpenCV convention and report the median rotation error, rotation accuracy, and translation error across all pairs. The rotation error is the minimum angular deviation between the predicted and the ground truth poses. In contrast, the translation error is the absolute difference between the corresponding translation vectors. We present accuracies as the percentage of pose  pairs with rotation errors below the thresholds of 15$^{\circ}$ and 30$^{\circ}$.
It should be noted that iFusion~\cite{wu2023ifusion} infers only the relative elevation, azimuth, and distance, and cannot provide the 4x4 camera matrix without the absolute camera pose of the reference image. For iFusion, we supplement the elevation angle of the reference image using an external elevation estimation method~\cite{liu2023one2345}, which has a median prediction error of 5$^{\circ}$ on the GSO dataset. Additionally, many baseline methods do not require camera intrinsics as input, resulting in predicted poses with varying distances from the camera to the shape, as reflected by the magnitude of the translation vectors. To address this intrinsic ambiguity, we normalize the predicted translation vectors for each method by using a scale factor that aligns the first view's predicted camera translation with the ground truth translation. After normalization, we report the absolute translation errors. Furthermore, in our ablation studies, we investigate the impact of the number of input views and the number of experts on pose estimation performance using subsets of 100 shapes.

\section{Details of Dataset Curation}
\label{sec:filter}
The Objaverse dataset~\cite{deitke2023objaverse} contains about 800,000 shapes. However, this dataset includes numerous partial scans, scenes, and basic, textureless geometries that are unsuitable for our task of generating single objects. To optimize the training process in terms of efficacy and efficiency, we curate a high-quality subset consisting of single objects with high-fidelity geometry and vivid textural appearance. We begin by randomly selecting a subset of 3D models and then task annotators with assessing the overall geometry quality and evaluating texture aesthetic preferences. Subsequently, we train a simple network to predict such annotations.

For assessing overall geometry quality, annotators are required to assign one of three possible levels to each 3D model:

\begin{quote}
    \textit{High quality}: Objects that represent a single entity with a clear semantic meaning, such as avatars and animals.

    \textit{Medium quality}: Simple geometric shapes (e.g., cubes, spheres); geometries that are abstract or have unclear semantic meaning; and repetitive structures found in the Objaverse, such as skeletal frames of houses and staircases.

    \textit{Low quality}: Point clouds; scenes with multiple elements; incomplete, low-quality, or unidentifiable 3D scans.
\end{quote}

For texture preference, given the difficulty in defining absolute standards due to aesthetic subjectivity, we adopt a binary choice approach for annotation. This method presents annotators with pairs of 3D models, prompting them to select the one with superior texture quality or visual appeal.

Overall, we have recruited 10 annotators and collected labels for 4,000 pairs of shapes in total. Based on these annotations, we trained MLP networks to predict overall geometry quality ratings and texture scores, respectively. Both networks take the multimodal features of each shape as input, which include image, text, and 3D features, as encoded in OpenShape~\cite{liu2024openshape}. The rating classification MLP predicts a one-hot encoded label across three levels, and is trained using the cross-entropy loss. Meanwhile, the texture scoring MLP regresses a score for each shape and is trained using a relative margin loss.

During the training of \ours, we utilized the trained MLPs to curate a subset of approximately 100,000 objects. These objects are rated as high-quality and possess texture scores within the top 20\%.

\end{document}